# A five-bar mechanism to assist fingers flexion-extension movement: System implementation

Araceli Zapatero-Gutiérrez*[1], Eduardo Castillo-Castañeda[1] and Med Amine Laribi[2]

[1]Centro de Investigación en Ciencia Aplicada y Tecnología Avanzada Unidad Querétaro, Instituto Politécnico Nacional, México.
[2]Institut PPRIME, Département Génie Mécanique et Systèmes Complexes, Université de Poitiers, France.
*Corresponding author.  E-mail: araceli_zapatero@hotmail.com.

## Abstract

The lack of specialized personnel and assistive technology to assist in rehabilitation therapies is one of the challenges facing the health sector today, a problem that is projected to increase in the coming years. For researchers and engineers, it represents an opportunity to innovate and develop devices that improve and optimize rehabilitation services for the benefit of society.
Hand injuries are classified within those that occur most frequently, and that will need a rehabilitation process to regain their functionality. This article presents the fabrication and instrumentation of an end-effector prototype for fingers rehabilitation that executes a natural flexion-extension movement. Based on a five-bar configuration, the dimensions were obtained through the gradient method optimization and evaluated thought Matlab. Experimental tests were carried out to demonstrate the prototype's functionality and effectiveness of a five-bar mechanism acting in a vertical plane.
A control position using $5^{th}$ order polynomials with via points was implemented in the joint space and the design of the end-effector was evaluated as a function of the angle of rotation; performing a theoretical comparison calculated as a function of a real flexion-extension trajectory of the fingers with the angle of rotation obtained through an IMU.

## 1. Introduction

The World Health Organization (WHO) notes that rehabilitation is an essential part of universal health coverage. It predicts that the need for rehabilitation will increase worldwide due to changes in the health and characteristics of the population. And it points out that in some low- and middle-income countries, more than 50% of people do not receive the rehabilitation services they need. In addition, it is an area that has also suffered the negative consequences of the COVID-19 pandemic, the rehabilitation services in 60-70% of countries have been affected [1].

The global rehabilitation needs remain unmet due to multiple factors, including the lack of qualified professionals to provide rehabilitation services; the ratio is less than ten qualified professionals per million inhabitants. And the lack of assistive technology and specialized equipment [1].

The development of assistive technology, such as rehabilitation devices, is a broad field of research and development, which, reaching the clinical stages, help the needs of physical rehabilitation.

The need for an individual to require physical rehabilitation can be due to a wide range of situations. Among the most frequently occurring injuries are hand injuries, which may require surgical and non-surgical medical attention. It should be noted that the hand represents one of the most important extremities of the upper limb, and mobility deficiencies that it may present have a direct impact on people's quality of life [2, 3]. Part of the recovery of the functionality of the hand involves the movement of the fingers, a common deficiency is the difficulty of the patient to extend the fingers [4].

The continuous flexion and extension movement of the fingers has proven to be a functional exercise for the recovery of the hand [5, 6].

One classification for robotic hand rehabilitation devices is through their interface with the user, divided into two categories, exoskeleton type and end effector type. Both categories have been shown to contribute substantially to the motor recovery of the fingers of the hand [3, 7-9].

The design of a finger rehabilitation mechanism that executes the flexion-extension movement was presented in [10]. The mechanism is designed as an end-effector device that executes a passive flexion-extension movement on the patient's fingers for the early stage of treatment. The device does not focus on performing specific tasks related to particular activities of daily life, such as grasping objects or fine motor function. This article presents the fabrication of the prototype presented in [10] and its instrumentation for laboratory testing.

The paper is structured as follows: Section 2 presents the conceptual proposal of the prototype design and the optimal dimensions founded by an optimization algorithm. Section 3 presents the development of the prototype including the fabrication and instrumentation process. Section 4 presents the prototype's end-effector trajectory and the comparison between the theorical rotation angle and the real rotation angle of the end-effector. Finally, in Section 5, the conclusions are presented.

## 2. Conceptual Prototype

The design of the prototype for rehabilitation of the flexion-extension movement of the fingers of the hand is based on the configuration of a five-bar mechanism. The five-bar mechanism has two degrees of freedom that allow it to generate different types of trajectories, a characteristic that makes it attractive for reproducing the natural flexion-extension movement of the fingers. The conceptual proposal of the design is explored in depth in [11], where the optimal dimensions of the prototype were found through a gradient method. The results obtained in the previous work are briefly retaken in this article.

### 2.1. Desired Trajectory

The desired trajectory for the end-effector of the five-bar mechanism corresponds to the natural flexion-extension movement of the fingers. As an end-effector mechanism, only the fingertip trajectory is considered.

From a group of healthy subjects, a set of representative curves of this movement were obtained. Through a Principal Component Analysis (PCA) it was possible to determine that the flexion-extension movement can be located within a plane, since there is minimal variation in one of the three coordinate axes. The representative curves for each finger vary in amplitude. The curve with the greatest amplitude (Figure 1), which corresponds to the movement of the middle finger, is considered the desired trajectory for this work [12].

Although only one curve is considered, the mechanism has been designed to modify the amplitude of the desired trajectory, as will be explained in section 3.

### 2.2. Five-bar Kinematic

The design of the rehabilitator is classified within the category of end-effector-type devices. Point $C$ of a five-bar symmetric mechanism generates a natural flexion-extension trajectory of the fingers, when the actuated joints $\theta_1$ and $\theta_2$ move from $\theta_{1_o}, \theta_{2_o}$ to $\theta_{1_f}, \theta_{2_f}$, respectively; as seen in Figure 2a.

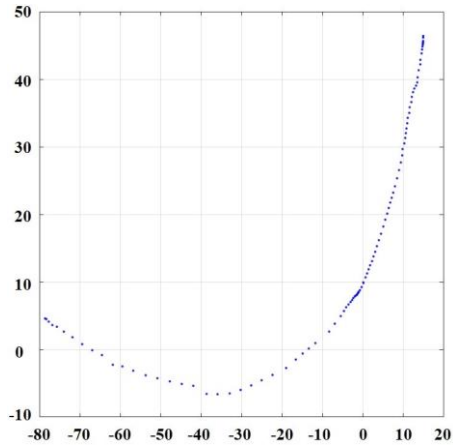

Figure 1. Real *flexion-extension trajectory.*

From Figure 2b, $L_1$, $L_2$, $L_3$, $L_4$ represent the lengths of the links, since $L_1 = L_4$ and $L_2 = L_3$ are a symmetric mechanism. The length $L_0$ is the distance between the fixed points $A$ and $E$. Point $C$ is a rotational joint that joins the links $L_2$ and $L_3$ [5], its kinematic equation is given by Equation (1) [11].

$$C = \begin{bmatrix} C_x \\ C_z \end{bmatrix} \quad (1)$$

Where

$$C_x = L_1 \cos\theta_1 + \frac{1}{2}[L_0 + L_1(\cos\theta_2 - \cos\theta_1)] - [L_1(\sin\theta_2 - \sin\theta_1)]\left[\sqrt{\frac{L_2^2}{H^2} - \frac{1}{4}}\right]$$

$$C_z = L_1 \sin\theta_1 + \frac{1}{2}[L_1(\sin\theta_2 - \sin\theta_1)] + [L_0 + L_1(\cos\theta_2 - \cos\theta_1)]\left[\sqrt{\frac{L_2^2}{H^2} - \frac{1}{4}}\right]$$

$$H^2 = L_0^2 + 2L_0L_1(\cos\theta_2 - \cos\theta_1) + 2L_1^2[1 - (\cos\theta_1 - \cos\theta_2)]$$

$\theta_1$ is computed from the sum of the angles $\beta$ and $\gamma$ as shown in Equation (2). While the difference between the $\omega$ and $\sigma$ angles compute $\theta_2$ given by Equation (3). The $\gamma$ and $\omega$ angles come from the real flexion-extension trajectory analyzed in [5] and shown in Figure 1 [11]. The five-bar mechanism point $C$ followa from $i$ point to $N$ point of the real trajectory expressed by D.

$$\theta_1 = \beta + \gamma \quad (2)$$
$$\theta_2 = \omega + \sigma \quad (3)$$

Where

$$\gamma = atan2(D_z, D_x)$$
$$\omega = \pi - atan2(D_z, L_0 - D_x)$$

To obtain the angles $\beta$ and $\sigma$, defined by Equations (4) and (5), it is necessary to calculate the $\varphi$ and $\alpha$ angles. The $\varphi$ and $\alpha$ angles are obtained by applying the law of cosines and the trigonometric properties of sine and cosine [10].

$$\beta = atan2(L_2 \sin(\pi - \varphi), L_1 + L_2 \cos(\pi - \varphi)) \quad (4)$$

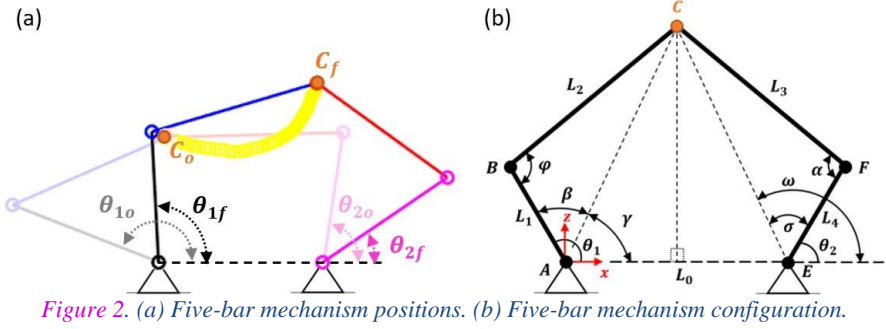

*Figure 2. (a) Five-bar mechanism positions. (b) Five-bar mechanism configuration.*

$$\sigma = atan2(L_3 \sin(\pi - \alpha), L_4 + L_3 \cos(\pi - \alpha)) \qquad (5)$$

Where

$$\varphi = atan2(\sin \varphi, \cos \varphi)$$

With

$$\cos \varphi = \frac{L_1^2 + L_2^2 - (D_x^2 + D_z^2)}{2L_1 L_2}$$

$$\sin \varphi = B\sqrt{1 - \cos \varphi^2}$$

And

$$\alpha = atan2(\sin \alpha, \cos \alpha)$$

With

$$\cos \alpha \frac{L_3^2 + L_4^2 - [(L_0 - D_{x_{i:N}})^2 + D_{z_{i:N}}^2]}{2L_3 L_4}$$

$$\sin \alpha = F\sqrt{1 - \cos \alpha^2}$$

There are four possible configurations for a five-bar mechanism, depending on the configuration of its elbows $B$ and $F$. The prototype presented in this document explores only one of the four configurations when elbows $B$ and $F$ are both up and they are represented by a magnitude of positive one [11].

### *2.3. Five-bar Optimal Solution*

The geometric parameters of the five-bar mechanism were obtained using an optimization algorithm based on the generation of the real flexion-extension trajectory (Figure 2). An objective function, given by Equation (6), minimize the error function $E(I)$ [11].

$$E(I) = \frac{1}{N}\sum_{i=1}^{N} \sqrt{(C_{x_i} - D_{x_i})^2 + (C_{z_i} - D_{z_i})^2} \qquad (6)$$

$E(I)$ is defined as the sum of the squared root of the $i_{th}$ position of the desired trajectory and the coordinates of point $C$, which represents the end-effector of the mechanism. Through the gradient method, implemented in MATLAB with the fmincon function, the optimal parameters of the design vector $I$, summarized in Table 1, were found.

The gradient method, also known as nonlinear programming, is used to find the minimum of a scalar function of several variables that begin with an initial estimate [13]. A maximum of 150 iterations with 2000 allowed evaluations of the function was chosen. The iterative process to minimize the objective function considers an upper and lower limit for each component of the design vector $I$. The design vector provides the length of the links and the initial position angles for $L_1$ and $L_4$ [11].

The optimization problem considers four constraint equations, Equations (7a), (7b), (7c), (7d), that ensure viable solutions for $H$, to guarantee the position of the end-effector, Equation (1), within the real numbers.

$$R_1 = \left|\frac{L_1^2 + L_2^2 - (D_{x_{i:N}}^2 + D_{z_{i:N}}^2)}{2L_1 L_2}\right| < 1 \tag{7a}$$

$$R_2 = \left|\frac{L_3^2 + L_4^2 - [(L_0 - D_{x_{i:N}})^2 + D_{z_{i:N}}^2]}{2L_3 L_4}\right| < 1 \tag{7b}$$

$$R_3 = |H_i| < 1 \tag{7c}$$

$$R_4 = \left|\sqrt{\frac{L_2^2}{H_i^2} - \frac{1}{4}}\right| < 1 \tag{7d}$$

## 3. Prototype Development

This section describes the development of the five-bar prototype based on the values of the optimal design vector $\boldsymbol{I}$. The purpose of the prototype is to evaluate the functionality of the design and check the tracking of the end-effector trajectory.

### 3.1. Design and Assembly

The prototype has two similar five-bar mechanisms that act in parallel. One of the mechanisms works as a master mechanism because it moves directly through the motors attached to its links $L_1$ and $L_4$. The second mechanism acts as a slave because it follows the movements of the master mechanism through guide bars and a connecting bar located at the junction of links $L_2$ and $L_3$, as shown in Figure 3. This bar also has the connections to adjust the length of the thimbles (Figure 4). A pair of brackets have been designed for the motors and the slave mechanism to be able to fix them to a base.

### 3.2. Materials, Components and Control

The links of the mechanism are made with polylactic acid (PLA) printed in 3D with 6 mm thick. The connecting bar between the master mechanism and the slave mechanism has a diameter of 7 mm; the connecting bar is designed to rotate freely to adapt to the position of the finger. Four 25 mm protruding bars are distributed along the connecting bar, which, in turn, are assembled with extension bars that allow adjusting the position of the thimbles (Figure 5).

The prototype movement is executed using two 6V DC motors (Pololu, 34:1 Metal Gearmotor with 48 CPR Encoder) controlled via a RoboClaw 7A (Motion Control) control card. The control interface was carried out using LabView (National Instruments) running on a PC (Intel Core i5-6300HQ, 8.00 GB RAM, 2.30 GHz).

*Table 1. Design vector I values.*

| *I[mm]* | | | *Input angles [degrees]* | | | |
|---|---|---|---|---|---|---|
| $L_1$ | $L_2$ | $L_0$ | $\theta_{1o}$ | $\theta_{1f}$ | $\theta_{2o}$ | $\theta_{2f}$ |
| 101.09 | 108.67 | 101.20 | 153.55 | 92.37 | 83.07 | 40.44 |

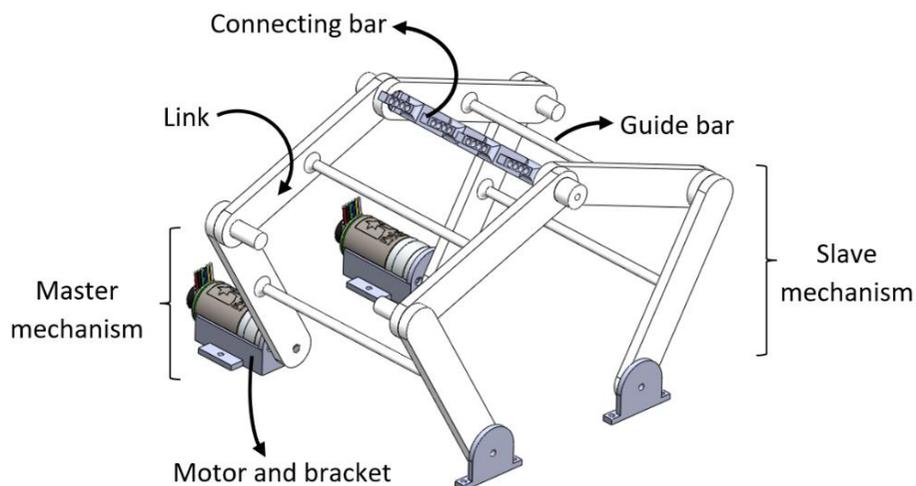

Figure 3. Rehabilitation prototype design.

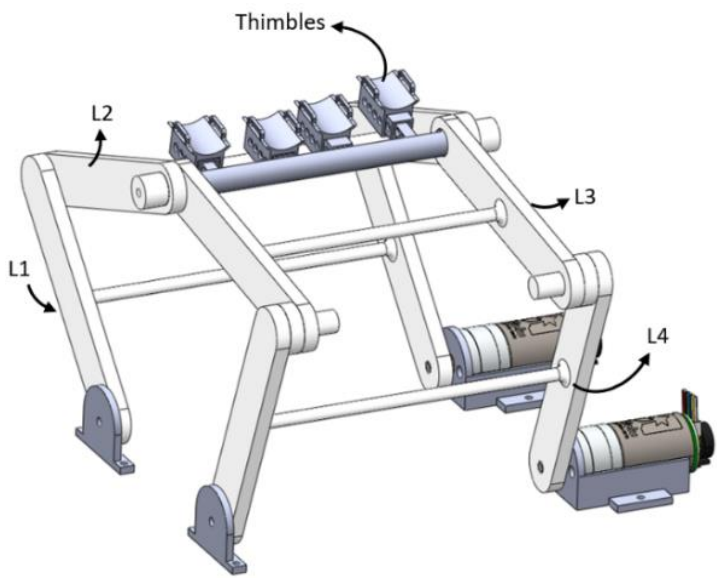

Figure 4. Rehabilitation prototype design, view with thimbles.

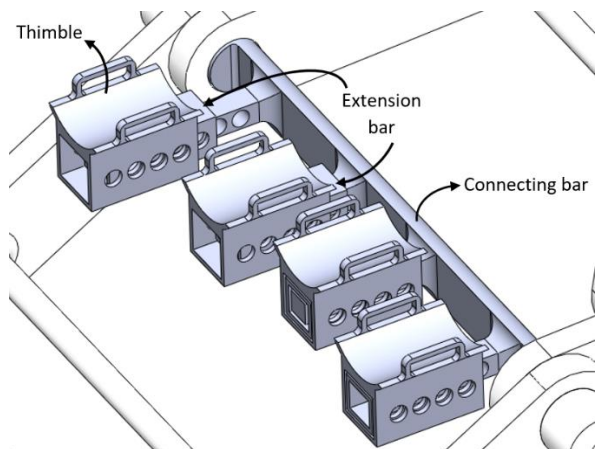

Figure 5. Connecting bar and thimbles adjustment.

The LabView program execute a point-to-point position control for the motors; a set of 5$^{th}$-order polynomial, given by the set of Equations (8a) to (8e), generates the trajectory in a period of ten seconds. From Equation (8a), $i$ takes the value 1 for the motor M1 and 2 for the motor M2. The joint trajectory considers zero initial and final values for the velocities and accelerations of both motors, as well two via points.

5$^{th}$-order polynomials were chosen for the trajectory in the joint space to ensure continuity in position, velocity and acceleration [14]. Equations (8e), (8i), (8m) and (8q) corresponds to the rate of change of acceleration (jerk); compute the jerk can assure a smooth movement and allows to obtain the coefficients of the polynomials.

The design of the rehabilitation mechanism considers that the movement of the fingers starts from the flexed finger position to the extended finger position. In this article, a *cycle* is defined as the finger trajectory from the flexion position to the extended position and vice-versa. In such a way that the trajectory begins and ends in the zero position as shown in Figure 6.

$$\theta_{Mi}(t) = \begin{cases} \theta_{1i}(t) & if\ 0 \leq t \leq t_{v1} \\ \theta_{2i}(t) & if\ t_{v1} \leq t \leq t_{f1} \\ \theta_{3i}(t) & if\ t_{f1} \leq t \leq t_{v2} \\ \theta_{4i}(t) & if\ t_{v2} \leq t \leq t_{f2} \end{cases} \tag{8a}$$

$$\theta_{1i}(t) = a_{01i} + a_{11i}t + a_{21i}t^2 + a_{31i}t^3 + a_{41i}t^4 + a_{51i}t^5 \tag{8b}$$

$$\dot{\theta}_{1i}(t) = a_{11i} + 2a_{21i}t + 3a_{31i}t^2 + 4a_{41i}t^3 + 5a_{51i}t^4 \tag{8c}$$

$$\ddot{\theta}_{1i}(t) = 2a_{21i} + 6a_{31i}t + 12a_{41i}t^2 + 20a_{51i}t^3 \tag{8d}$$

$$\dddot{\theta}_{1i}(t) = 6a_{31i} + 24a_{41i}t + 60a_{51i}t^2 \tag{8e}$$

$$\theta_{2i}(t) = a_{02i} + a_{12i}t + a_{22i}t^2 + a_{32i}t^3 + a_{42i}t^4 + a_{52i}t^5 \tag{8f}$$

$$\dot{\theta}_{2i}(t) = a_{12i} + 2a_{22i}t + 3a_{32i}t^2 + 4a_{42i}t^3 + 5a_{52i}t^4 \tag{8g}$$

$$\ddot{\theta}_{2i}(t) = 2a_{22i} + 6a_{32i}t + 12a_{42i}t^2 + 20a_{52i}t^4 \tag{8h}$$

$$\dddot{\theta}_{2i}(t) = 6a_{32i} + 24a_{42i}t + 60a_{52i}t^2 \tag{8i}$$

$$\theta_{3i}(t) = b_{01i} + b_{11i}t + b_{21i}t^2 + b_{31i}t^3 + b_{41i}t^4 + b_{51i}t^5 \tag{8j}$$

$$\dot{\theta}_{3i}(t) = b_{11i} + 2b_{21i}t + 3b_{31i}t^2 + 4b_{41i}t^3 + 5b_{51i}t^4 \tag{8k}$$

$$\ddot{\theta}_{3i}(t) = 2b_{21i} + 6b_{31i}t + 12b_{41i}t^2 + 20b_{51i}t^3 \tag{8l}$$

$$\dddot{\theta}_{3i}(t) = 6b_{31i} + 24b_{41i}t + 60b_{51i}t^2 \tag{8m}$$

$$\theta_{4i}(t) = b_{02i} + b_{12i}t + b_{22i}t^2 + b_{32i}t^3 + b_{42i}t^4 + b_{52i}t^5 \tag{8n}$$

$$\dot{\theta}_{4i}(t) = b_{12i} + 2b_{22i}t + 3b_{32i}t^2 + 4b_{42i}t^3 + 5b_{52i}t^4 \tag{8o}$$

$$\ddot{\theta}_{4i}(t) = 2b_{22i} + 6b_{32i}t + 12b_{42i}t^2 + 20b_{52i}t^3 \tag{8p}$$

$$\dddot{\theta}_{4i}(t) = 6b_{32i} + 24b_{42i}t + 60b_{52i}t^2 \tag{8q}$$

Considering that:

$$\theta_{1i}(t_{v1}) = \theta_{2i}(t_{v1}) \tag{8r}$$

$$\dot{\theta}_{1i}(t_{v1}) = \dot{\theta}_{2i}(t_{v1}) \tag{8s}$$

$$\ddot{\theta}_{1i}(t_{v1}) = \ddot{\theta}_{2i}(t_{v1}) \qquad (8t)$$

$$\theta_{3i}(t_{v2}) = \theta_{4i}(t_{v2}) \qquad (8u)$$

$$\dot{\theta}_{3i}(t_{v2}) = \dot{\theta}_{4i}(t_{v2}) \qquad (8v)$$

$$\ddot{\theta}_{3i}(t_{v2}) = \ddot{\theta}_{4i}(t_{v2}) \qquad (8w)$$

The polynomial given by Equation (8b), provides the flexion (zero) position to the first via point; the polynomial given by Equation (8f), provides the position from the first via point to the extension (final) position of the finger. $\theta_{1i}$ and $\theta_{2i}$ correspond to the half of a cycle.

Equation (8j) express the movement from the extended position to the second via point. Finally, Equation (8n), provides the position from the second via point to flexion position of the finger, completing a cycle.

The coefficients of the polynomials corresponding to the first half of the cycle ($C_{fe}$) are calculated through Equation (9a), multiplying the inverse matrix $Mz_{1i}$ that considers equations (8b) to (8i) by the vector of defined conditions $q_{1i}$.

$$[C_{fe}]_{12\times1} = [Mz_{1i}]^{-1}{}_{12\times12}\,[q_{1i}]_{12\times1} \qquad (9a)$$

Where:

$$q_{1i} = \begin{bmatrix} \theta_{1i}(0) \\ \dot{\theta}_{1i}(0) \\ \ddot{\theta}_{1i}(0) \\ \dddot{\theta}_{1i}(0) \\ \theta_{1i}(t_{v1}) \\ \theta_{2i}(t_{v1}) \\ 0 \\ 0 \\ \theta_{2i}(t_{f1}) \\ \dot{\theta}_{2i}(t_{f1}) \\ \dddot{\theta}_{2i}(t_{f1}) \\ \ddot{\theta}_{2i}(t_{f1}) \end{bmatrix}$$

The coefficients of the polynomials corresponding to the second half of the cycle ($C_{ef}$) are calculated through Equation (9b), multiplying the inverse matrix $Mz_{2i}$ that considers equations (8j) to (8q) by the vector of defined conditions $q_{2i}$.

$$[C_{ef}]_{12\times1} = [Mz_{2i}]^{-1}{}_{12\times12}\,[q_{2i}]_{12\times1} \qquad (9b)$$

Where:

$$q_{2i} = \begin{bmatrix} \theta_{3i}(t_{f1}) \\ \dot{\theta}_{3i}(t_{f1}) \\ \dddot{\theta}_{3i}(t_{f1}) \\ \ddot{\theta}_{3i}(t_{f1}) \\ \theta_{3i}(t_{v2}) \\ \theta_{4i}(t_{v2}) \\ 0 \\ 0 \\ \theta_{4i}(t_{f2}) \\ \dot{\theta}_{4i}(t_{f2}) \\ \dddot{\theta}_{4i}(t_{f2}) \\ \ddot{\theta}_{4i}(t_{f2}) \end{bmatrix}$$

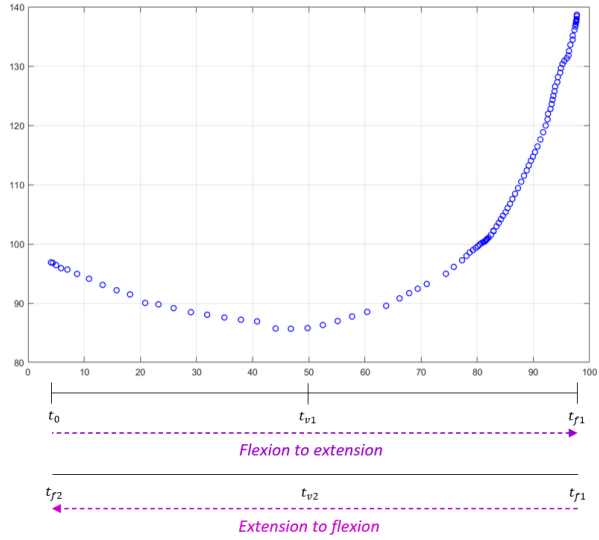

Figure 6. A cycle of movement.

An MPU-6050 IMU was placed on the connecting bar to measure the end-effector angle on the coordinate axes through an Arduino UNO. The connection is independent of the motor control program. The prototype instrumentation shown in Figure 7.

### 3.3. Experimental evaluation

The flexion-extension trajectory runs in ten seconds and represents one cycle of motion. The encoders were initialized to zero for 153° in the $L_1$ link connected to the $M_2$ motor, and 83° in the $L_4$ link connected to the $M_1$ motor, approximately, as shown in Figure 8. The encoder resolution is 48.14 counts for one turn at the out of the reducer. Table 2 indicates the relationship between the articular positions and the positions of the actuated links, at the initial, vias, and final points; obtained through inverse kinematics. The via points was computed for $t_{v1} = 2.5\ s$ and $t_{v2} = 7.5\ s$. Figure 9 shows the profiles of the flexion-extension trajectory of both motors in radians.

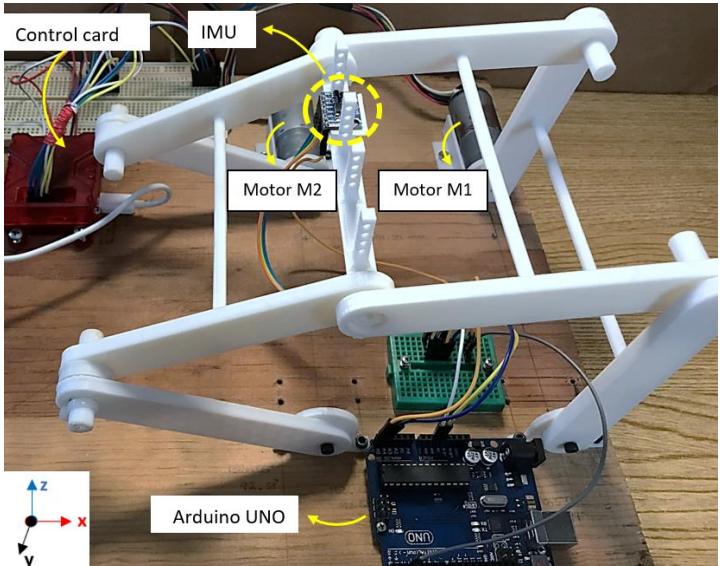

Figure 7. Mechanism instrumentation.

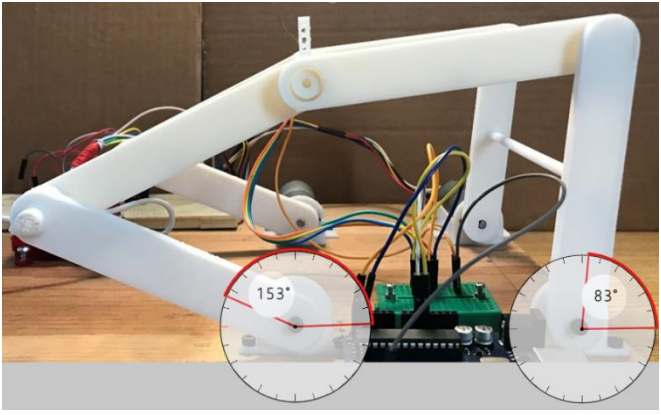

*Figure 8. L1 and L4 links orientation.*

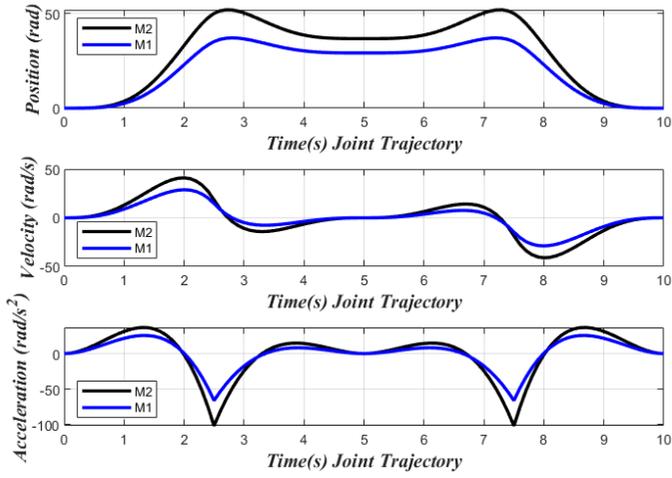

*Figure 9. Joint movement profiles.*

*Table 2. Position setting.*

| Positions | Articular (encoder) | Links (degrees) |
|---|---|---|
| $\theta_{M1}(0)$ | 0 | 83.07 |
| $\theta_{M1}(2.5)$ | 201 | 35.99 |
| $\theta_{M1}(5)$ | 155 | 40.44 |
| $\theta_{M1}(7.5)$ | 201 | 35.99 |
| $\theta_{M1}(10)$ | 0 | 83.07 |
| $\theta_{M2}(0)$ | 0 | 153.55 |
| $\theta_{M2}(2.5)$ | 379 | 104.39 |
| $\theta_{M2}(5)$ | 277 | 92.37 |
| $\theta_{M2}(7.5)$ | 379 | 104.39 |
| $\theta_{M2}(10)$ | 0 | 153.55 |

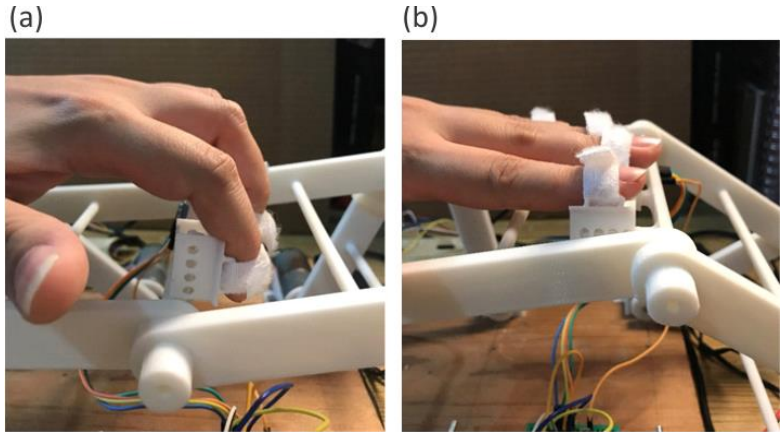

*Figure 10. (a) Finger flexed position. (b) Finger extended position.*

The IMU is directly coupled to the connecting bar and measures the rotation of the bar with respect to the *x*, *y*, and *z* axes. The connecting bar rotates freely and adjusts to the position of the finger as the mechanism performs the movement, as shown in Figure 10.

The angles obtained by the IMU in the *x*-axis ($\zeta_x$), the *y*-axis ($\zeta_y$) and the *z*-axis ($\zeta_z$) during the four cycles are shown in Figure 11. The greatest dispersion is found in the *y*-axis due to the rotational joint that joins the $L_2$ and $L_3$ links, and that is directly associated with the connecting bar. The *x* and *z* axes show little dispersion because it is a planar mechanism. It was observed that the connecting bar presents a little friction when rotating almost to the end of the trajectory.

## 4. Results

Figure 12 shows a sequence of images the different positions of the mechanism considering the via points. While Figure 13a shows the interpolation generated by these points and the estimated trajectory they generate.

Considering the translational movement in the *xz* plane of the connecting bar, the angle of rotation obtained in the *y*-axis through the IMU, can be compared with the theoretical angle. The theoretical angle was obtained through Matlab. The inverse and direct kinematics of the mechanism are calculated using the optimal results of the gradient method (Section 2) [10], and the points generated by the desired trajectory of Figure 1, considering a flexion-extension cycle.

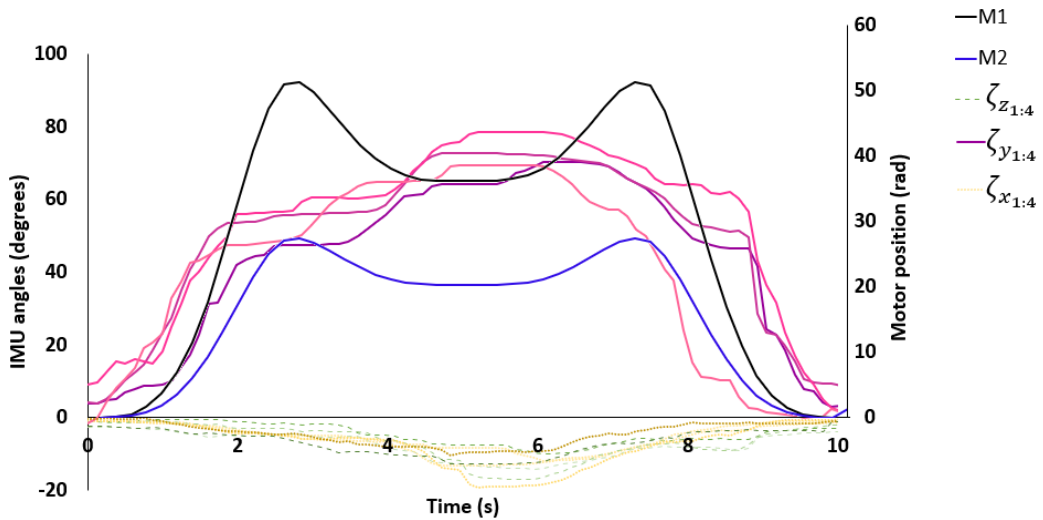

*Figure 11. Comparison between the motor position and rotation angles obtained by the IMU.*

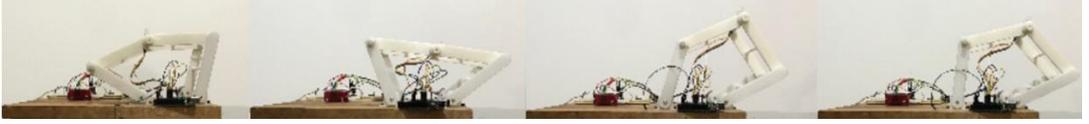

*Figure 12. Diferent positions of the mechanism.*

With the coordinates of point $C$ $(C_x, C_z)$, a new vector $V_{pi}$ is defined, $V_{pi}$ change from the initial position from the $i^{th}$ position. The $V_p$ vector represents the relative position of two points, expressed by equation (10). The $V_p$ vector joins two continuous points of the trajectory generated by the end-effector; the rectangular components $V_r$ and $V_p$ allow to calculate the theoretical angle $\zeta_{yTi}$ of the trajectory points through equation (11) according to the variation of the position, as shown in Figure 13b.

$$V_p = (C_{xi} - C_{xi-1})\hat{\imath} + (C_{zi} - C_{zi-1})\hat{k} = V_r\hat{\imath} + V_p\hat{k} \tag{10}$$

$$\zeta_{yT} = \operatorname{atan}\left|\frac{V_p}{V_r}\right| \tag{11}$$

The $\zeta_{yT}$ was computed only considering the movement from the flexion to extension position to not overwrite trajectory points. Figure 14a shows the theoretical angle obtained; in Figure 14b the comparison among the theoretical angle with respect to the four cycles obtained with the IMU can be observed. For the purpose of comparison, an offset has been applied to each curve so that they start at zero at time zero. Figure 14c shows the error of each cycle with respect to the theoretical angle, the friction presented by the connecting bar is directly related to the error.

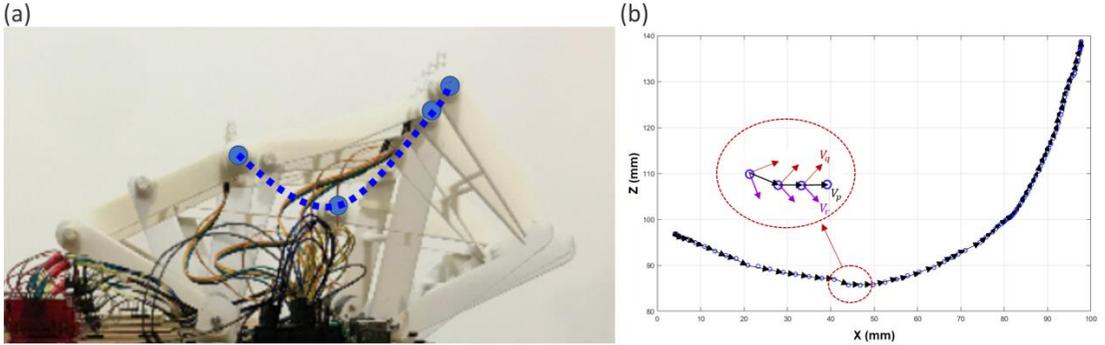

*Figure 13. (a) Sequence of positions. (b) $V_p$ vector.*

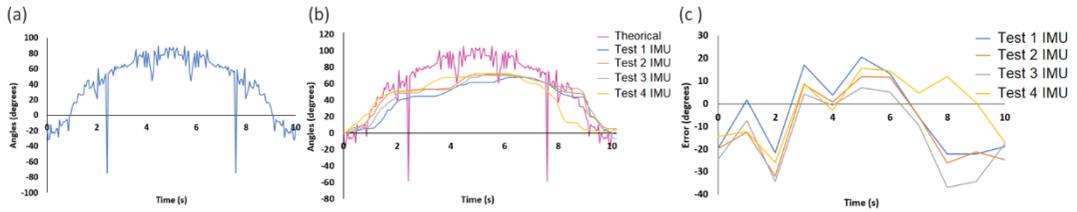

*Figure 14. (a) $\zeta_{yT}$ versus the four $\zeta_y$ obtained in the four cycles. (b) Error by cycle.*

## 5. Conclusions

The manufacture and validation of an end-effector type prototype for the rehabilitation of fingers of the hand was presented. The mechanism is based on the configuration of a mechanism with five bars and two degrees of freedom. Manufactured in 3D printing and controlled in the joint space by fifth order polynomials.

Regarding the design of the prototype, it was observed that several points of improvement that will be explored in the evolution of the project. Gravity is a factor that influences movement, although the mechanism is capable of following the points of the trajectory in the plane; the choice of the *xz* plane represents considering the effects of the weight of the links to keep the elbows (points *B* and *D*) up; it was observed that after several tests the desired position tends to be lost due to the elbows falling.

As immediate future work is to improve the rotational articulation that carries the connecting bar; placing a bearing in this part can solve the friction problem that influences the measurement of the angle of rotation.

Future work will also consider the study and analysis of materials applicable to medical device regulations, as well as the development of a mechanism-patient interaction control algorithm.